# Predictive Situation Awareness for Ebola Virus Disease using a Collective Intelligence Multi-Model Integration Platform: Bayes Cloud[1]

*Cheol Young Park[a]\*, Shou Matsumoto[b], Jubyung Ha[a], YoungWon Park[a]*

[a]BAIES, LLC, USA. [b] George Mason University, USA.

ABSTRACT

The humanity has been facing a plethora of challenges associated with infectious diseases (e.g., Zika, MERS, Ebola, SARS and H1N1), which kill more than 6 million people a year. Although continuous efforts have been applied to relieve the potential damages from such misfortunate events, it's unquestionable that there are many persisting challenges yet to overcome. One related issue we particularly address here is the assessment and prediction of such epidemics. In this field of study, traditional and/or ad-hoc models frequently fail to provide proper predictive situation awareness, characterized by understanding the current situations and predicting the future situations. Comprehensive predictive situation awareness for infectious disease can support decision making and help to hinder disease spread. In this paper, we develop a computing system platform focusing on collective intelligence causal modeling, in order to support predictive situation awareness in the domain of infectious disease. Analyses of global epidemics require integration of multiple different data and models, which can be originated from multiple independent researchers. These models should be integrated to accurately assess and predict the infectious disease in terms of holistic view. The system shall provide three main functions: (1) collaborative causal modeling, (2) causal model integration, and (3) causal model reasoning. These functions are supported by subject-matter expert (SME) and artificial intelligence (AI), with uncertainty treatment. Subject-matter experts, as collective intelligence, develop causal models and integrate them as one joint causal model. Artificial intelligence is used to find causal models from texts, by using natural language processing. The integrated causal model shall be used to reason about: (1) the past, regarding how the causal factors have occurred; (2) the present, regarding how the spread is going now; and (3) the future, regarding how it will proceed. Finally, we introduce one use case of predictive situation awareness for the Ebola virus disease.

## 1. Introduction

The humanity has been facing a plethora of challenges associated with infectious diseases (e.g., Zika, MERS, Ebola, SARS and H1N1), which killed more than 6 million people in 2016 only [BCM, 2018]. Infectious diseases are a kind of diseases caused by micro-sized organisms (e.g., bacteria and viruses). The diseases spread out through direct and indirect contacts between animals and humans. Historically, such diseases developed rapidly and caused a lot of fatalities. As can be easily expected, the outbreak of the diseases is ongoing world-wide currently as well as will be present in the future. Although continuous efforts have been applied to relieve

---

the potential damages from such misfortunate events, it's unquestionable that there are many persisting challenges yet to overcome. One approach we particularly address in this project is the awareness and prediction of such epidemic outbreaks to prevent the wide spread of the diseases. In this field of study, traditional and/or ad-hoc models are used to focus on only one aspect of each independent domain (e.g., prediction of virus mutation, analysis between economic situation and disease occurrence, and analysis for patient disease status). However, these independent models fail to provide proper predictive situation awareness, characterized by understanding current situations and predicting future situations. Comprehensive predictive situation awareness for infectious disease can support decision making and help to hinder infectious disease spread. In this paper, we introduce a computing system platform focusing on a *Collective Intelligence Multi-Model Integration Platform*, called *Bayes Cloud*, in order to support predictive situation awareness in the domain of infectious disease.

Bayes Cloud provides three main functions: (1) collaborative causal modeling, (2) causal model integration, and (3) causal model reasoning. The three functions are supported by SME and/or AI, with uncertainty treatment for the associated sample spaces. Analyses of global epidemics require integration of multiple different data and models, which are originated from multiple independent researchers. If these models can be integrated as one model, it can be used to analyze the epidemic situation in terms of holistic view. Bayes Cloud, we introduce in this paper, aims to integrate independent models into one collective model. The current version of Bayes Cloud can deal with causal models [Pearl & Mackenzie, 2018] and Bayesian Network (BN) models [Pearl, 1988]. The future version will embody Multi-Entity Bayesian Network (MEBN) models [Laskey, 2008]. Subject-matter experts, as collective intelligence, develop causal models and integrate them as one joint causal model (a.k.a. model integration). Model integration can be viewed as a process for merging several models in order to obtain a third model. The resulting model shall embody the information from the original models, and also support joint reasoning unattainable solely by the individual models. This integration process can be performed either manually or automatically using Bayes Cloud. In this paper, we introduce automatic model integration. Natural Language Processing (NLP) can be used to find causal models from the texts concerning a specific epidemic disease domain. Although this approach does not guarantee perfect results, as a prior knowledge for the causal models, the results can be useful to SME or model developers. The integrated causal model is used to reason about the occurrence in: (1) the past, regarding how the causal factors have occurred; (2) the present, regarding how the spread is going now; and (3) the future, regarding how it will proceed. In this paper, we introduce how to apply Bayes Cloud to the epidemic disease domain to support predictive situation awareness. Particularly, we treat the case of the Ebola disease.

In Section 2, background knowledge is given first. Brief definition for BN is introduced. Next, the recent researches for the Ebola virus disease are introduced. In Section 3, the collective intelligence multi-model integration platform, Bayes Cloud, is introduced. In Section 4, the predictive situation awareness model for the Ebola virus disease using Bayes Cloud is introduced. Finally, conclusions are presented and future research directions are discussed.

## 2. Background

In this section, we describe the definition of BN and the current researches for the Ebola virus disease.

### 2.1. Bayesian Network

A Bayesian Network (BN) [Pearl, 1988] is a probabilistic graphical model that represents a joint distribution on a set of random variables in a compact form that exploits conditional independence relationships. Random variables (RVs) are represented as nodes in a directed acyclic graph (DAG) in which a directed edge represents a direct dependency between two nodes and no directed cycles are allowed. We introduce the following definition for BN.

**Definition 2.1** (Bayesian Network) A Bayesian Network (BN) for a set of RVs X = {$X_1$, $X_2$, …, $X_n$} is a pair [G, Θ], where G is a DAG whose nodes are associated with the RVs and Θ = {$θ_1$, …, $θ_n$} is a set of local distributions, where $θ_i$ = P($X_i$ | Pa($X_i$)) is the conditional distribution of $X_i$ given its parents Pa($X_i$) in G. A Bayesian network represents the joint distribution of a set of RVs X as a product of the conditional distributions of $X_i$ given its parents Pa($X_i$) in G:

$$P(X_1, X_2, \ldots, X_n) = \prod_{i=1}^{n} P(X_i | Pa(X_i)).$$

Fig. 1 shows the simple BN for the Ebola Virus Disease (EVD) that represents an illustrative example for the EVD diagnosis. This BN contains two nodes (*EbolaVirusDisease* and *Haemorrhage*) and an arc, and local distributions.

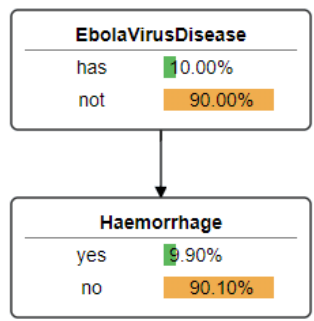

**Figure 1. A simple discrete BN for the Ebola Virus Disease**

The random variable (RV) or node Haemorrhage is used to represent some observed findings (e.g., symptoms and diagnosis results) for a patient. The node EbolaVirusDisease is used to predict the existence of a certain disease (e.g., has the Ebola virus disease) for the patient. As shown by the arc direction in the figure, EbolaVirusDisease influences Haemorrhage.

The BN can be described by a Bayesian Network Script as shown in Script 1. The Bayesian Network Script defines the nodes (EbolaVirusDisease and Haemorrhage) and their local distributions.

**[Script 1] Discrete Bayesian Network Script for the simple Ebola virus disease BN**

```
1    defineNode(EbolaVirusDisease, Description);
2    {
3        defineState(Discrete, has, not);
4        p(EbolaVirusDisease) =
5            {has: 0.1; not: 0.9;}
6    }
7
8    defineNode(Haemorrhage, Description);
9    {
10       defineState(Discrete, yes, no);
11       p( Haemorrhage | EbolaVirusDisease) =
12           if (EbolaVirusDisease == has)
13               {yes: 0.9; no: 0.1;}
```

```
14          else if (EbolaVirusDisease == not)
15              {yes: 0.01; no: 0.99;}
16      }
```

In Line 1, the node EbolaVirusDisease is defined. In Line 3, the type of states in EbolaVirusDisease is defined as discrete and its two states (i.e., *has* (the disease) and *not* (have the disease)) are specified. In Line 5, probability values for the both of these states are specified. From Line 8 to Line 16, the node Haemorrhage is defined. The conditional probability distributions for this node are specified by using if-statement. For example, the symptom Haemorrhage happens (i.e., the state *yes*) with the probability of 90%, if a patient has the Ebola virus disease (i.e., the state *has*).

Figure 1 shows a discrete BN, while Figure 2 represents a hybrid BN, which contains both discrete and continuous random variables (or nodes). The continuous nodes can be conditional linear/nonlinear probability distributions (e.g., normal and beta distribution). The following equation illustrates the conditional linear Gaussian (CLG) distribution.

$$p(N \mid \text{Pa}(N), CF_j) = \mathcal{N}(m + b_1 P_1 + b_2 P_2 \ldots, +b_n P_n, \sigma^2),$$

where Pa() is a set of continuous parent resident nodes of the continuous resident node, N, having $\{P_1, \ldots, P_n\}$, $CF_j$ is a *j*-th configuration of the discrete parents of *N*, m is a regression intercept, $\sigma^2$ is a conditional variance, and $b_i$ is regression coefficient.

Figure 2 shows the graphical version of a hybrid BN. For example, there are the parent node *EbolaVirusDisease* with 2 configurations (*has* and *not*) and the child node *Fever*, which is a continuous node represented by a mixture Gaussian distribution. If a patient does not have EVD, the temperature of the patient is about 98.6 F, while if the patient has EVD, the patient accompanies a high fever of 103 F. The same contents of Figure 2 are written as a BN Script as shown in Script 2.

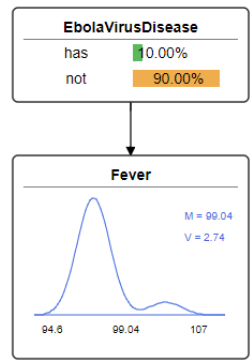

**Figure 2. a simple hybrid BN**

**[Script 2] Hybrid Bayesian Network Script**

```
1   defineNode(EbolaVirusDisease, Description);
2   {
3       defineState(Discrete, has, not);
4       p(EbolaVirusDisease) =
5           {has: 0.1; not: 0.9;}
6   }
```

```
7
8    defineNode(Fever, Description);
9    {
10       defineState(Continuous);
11       p(Fever | EbolaVirusDisease) =
12          if (EbolaVirusDisease == has)
13             { NormalDist(103, 1.0) }
14          else if (EbolaVirusDisease == not)
15             { NormalDist(98.6, 1.0) }
16   }
```

In Line 12 and 14 at Script 2, there are If-else nested statements. By these, the normal distributions are assigned to the two states (i.e., *has* and *not*) of the parent node EbolaVirusDisease. For example, Line 12 denotes that the temperature of an EVD patient is a normally distributed with the mean temperature of 103 F and with the variance of 1.0.

BN is a powerful tool for representing uncertain knowledge and performing inference under uncertainty. BN has been applied in a wide range of medical domains, which are titled medical decision making. Cooper [1984] classified medical decision making using BN into three categories of *diagnosis*, *treatment*, and *prognosis*. Lucas et al. [2004] introduced four categories: *diagnostic reasoning*, *prognostic reasoning*, *treatment selection*, and *discovering functional interactions*.

Diagnostic reasoning is a task to find a patient's disease using evidence or findings. Examples of such evidence include patient history, information, and diagnostic test results. In addition to existing patient data, additional testing is required. Because the performances (e.g., accuracy and precision) of the additional testing results are very different, selection for additional testing is considered another important issue in diagnosis. For these reasons, BNs for diagnosis often contain various random variables such as patient symptom, history, condition, and testing results. Prognostic reasoning is about prediction for a patient status. Disease and condition status of a patient will change over time. Understanding such change helps medical doctors to perform better medical decision making. Dynamic or temporal BNs (e.g., [Murphy, 2002][Laskey, 2008]), which represent temporal aspects, can be used for prognostic reasoning. We can also choose appropriate treatments and additional testing through prognostic reasoning. Treatment selection is a task to choose proper treatments for a patient. Treatment selection is performed by the subjective decision making of medical doctors in conjunction with a patient. BN allows such subjective decisions to be integrated into existing BN models. Discovering functional interactions is to analyze relationships among factors involved. BN can represent causality and correlation among variables. Such explicit knowledge enables us to perform better medical analysis.

Cooper [1984] researched the domain of hypercalcemic disorders to enable physicians to make decision for the diagnostic hypothesis depending on patient's evidences. BN was used for the research regarding the diagnosis of muscle and nerve diseases using electromyography [Andreassen et al., 1987]. Velikova et al. [2014] introduced a temporal BN model for pregnancy care. Prediction for glioblastoma multiforme using BN was researched by integrating a BN from several existing statistical models [Singleton, 2016]. The diagnosis of lymph-node diseases to support surgical pathologists was researched by [Heckerman et al., 1992]. The diagnosis of heart diseases using BN was researched by [Diez et al., 1997][Zarandi et al., 2017].

Fenton & Neil [2010] introduced the advantages of BN in medical negligence case. BN has three characteristics: (1) *Human-understandable model*, it provides a clear idea of causal relations between factors without distortion, (2) *Integrable model*, it can be represented as one model for separate medical pathways (cf. other machine learning models, such as decision tree, may require several separate decision tree models to represent one complex situation), and (3) *Assumption-applicable model*, it can be used to represent different scenarios by changing different prior probability assumptions.

*2.2. Ebola Disease*

The Ebola Virus Disease (EVD) outbreak seems to have emerged in December 2013 and identified March 2014 spreading around the sub-Saharan African region. The EVD outbreak had a high fatality rate of 30 to 90% depending on virus species [Baize et al., 2014] and was studied by researchers around the world. World Health Organization (WHO) estimated the EVD reproduction for three countries (Guinea, Liberia, and Sierra Leone) and predicted a rapid increase of the cases in the absence of control [WHO Ebola Response Team, 2014]. Carroll et al. [2015] assumed that the origin of the virus was a zoonotic transmission from bats to a two-year-old boy in December 2013, and then spread through human contact throughout the sub-Saharan African region. By December 2015 the number of infected individuals stood at 28640 which resulted in 11315 fatalities (39.5% fatality rate) [Ngwa & Teboh-Ewungkem, 2016]. The symptoms of EVD includes fever, fatigue, loss of appetite, vomiting, diarrhoea, headache, abdominal pain, muscle pain, joint pain, chest pain, cough, difficulty breathing, difficulty swallowing, conjunctivitis, sore throat, confusion, hiccups, jaundice, eye pain, rash, coma, and unconsciousness [WHO Ebola Response Team, 2014].

*2.2.1. Epidemiological Analysis for EVD*

Various types of EVD epidemiological analysis have been researched to understand the characteristics of EVD, predict dynamics, and identify optimal control strategies (e.g., vaccination and containment). Such analysis may be able to decelerate the EVD spread and prevent wide spread of future epidemics.

The basic research of EVD epidemiological analysis includes a basic reproduction number ($R_0$), which is the average number of secondary cases that occur when an initial case is observed. If $R_0$ is less than one, the EVD spread decreases (i.e., Eradication), while if $R_0$ is higher than one, the EVD spread increases (i.e., Epidemic). Such the basic reproduction number can be influenced by various control factors (e.g., environmental, social, temporal, and geospatial factors). Finding or estimating strongly related factors have been one of the active research themes. The aspects of the EVD spread appear through four stages: (1) *Incubation period*, (2) *Prodromal period*, (3) *Infectious period*, and (4) *Recovery period*. Researchers have studied the dynamics for infectious patients over these stages. Estimating serial interval, the time between the primary case and the secondary case for EVD onset, is also crucial to decrease the spreading [Fine, 2003][Wallinga & Lipsitch, 2007].

EVD spreads in many ways. The patterns for transmission between human and reservoirs also have been actively researched (e.g., [Rivers et al., 2014][Van Kerkhove et al., 2015][Carroll et al., 2015]). While the zootomic source is believed to have originated in fruit bats, there have been instances of other non-human primates and forest antelopes becoming infected and being vectors of transmission. Because of this, it is believed that most EVD outbreaks begin with an animal reservoir which then transitions into human-to-human transmission [Van Kerkhove et al., 2015]. While direct transmission assures infection, there was the indirect transmission. Berge et al. [2017] introduced how EVD spreads even without direct contact with infected individuals. Reduction for transmission was also researched. A combination of increased contact tracing and improved infection control would have a vast impact on the number of Ebola case reducing the number of people infected [Rivers et al., 2014]. While travel bans such as air travel or border control help mitigate the infection of Ebola, it does not completely prevent and only postpones it by weeks or months depending on the distance [Poletto et al., 2014]. However, these measures can be harmful to the country and/or nation in question in the long run causing reduced humanitarian efforts and economic problems as well. EVD was highly contagious spreading via bodily fluids and direct contact, and also indirect contact even spread by air as in the case of Nigeria. Browne et al. [2014] pointed out the importance of containing Ebola virus through mathematical tracing to both identify and isolate the infected individuals to contain the outbreak. Lau et al. [2017] introduced an age-specific spatiotemporal statistical framework for EBOV. In the research, the transmission dynamics and potential drivers of spreading were identified and applied to the framework. The framework was used to reason about the probability distribution of the number of the new patient cases for each infected individual.

A SEIR (susceptible, exposed, infectious, and recovered) model is used to represent an infectious patient status in epidemic dynamics. For EVD epidemic dynamics, a variety of SEIR models aimed to represent reproduction of the Ebola virus is introduced by several researches (e.g., [Althaus, 2014][Chowell & Nishiura, 2014] [Scarpinoet al., 2014][Browneet al., 2014]). Such models can be used for measuring the disease spread and status of current public health. Cleaton et al. [2015] indentified the key factors for transmission pattern in the 2014-2015 Ebola outbreak by taking news reports. Lewnard et al. [2014] researched models for intervention strategies controlling the spread of Ebola virus. Their models were used to estimate the effects of Ebola virus disease treatment centres, case ascertainment through contact tracing, and household protective kit allocation. Other researchers have made other attempts in tracing the virus by modelling using virus age where infected individuals acquire the disease where the virus incubates from 2-21 days. After virus incubation the individual can be removed from the population either by hospitalization or mortality, helping to mitigate the epidemic [Webb & Browne, 2016].

Grassly & Fraser [2008] classified three infectiousness types: (1) *Biological infectiousness*, (2) *Behavioural infectiousness*, and (3) *Environmental infectiousness*. Biological infectiousness depends on the biological characteristics between the pathogen and the host. For example, the efficiency of the host immune system can affect infectivity. Behavioural infectiousness is about the contact patterns of infected individuals. Environmental infectiousness is associated with the location and environmental factors for the infected individuals. In this view, three different models can be considered to build the epidemiological model of the EVD spread. These infectiousness types can be the subject of analysis and model development.

### 2.2.2. Mathematical Models for EVD

Various types of mathematical models have been used for EVD epidemiological analysis. The mathematical models for epidemiological analysis can be classified into deterministic models and stochastic models. The deterministic models may consist of a schematic relationship, representing a causal relationship among factors, and deterministic math formulas among these factors. The deterministic models are used to predict outputs of the factors (e.g., the number of susceptible and infectious patients) over time (i.e., dynamics). The stochastic models can deal with the dynamics as well as the uncertainty. The stochastic models may contain a causal relationship, but are based on probability theory. The main advantage of the stochastic models is high prediction accuracy, however the computation of such models is slower than the deterministic models and also building such models can be cumbersome for a complex case.

Siettos & Russo [2013] introduced three classes of models for infectious disease dynamics: (1) *Statistical-Based Methods*, (2) *Mathematical/Mechanistic State-Space Models*, and (3) *Empirical/Machine Learning-Based Models*. Statistical-Based Methods include regression models, time series analysis, statistical process control methods (e.g., cumulative sum chart), Hidden Markov models, and spatial models (for spatio-temporal pattern). Mathematical/Mechanistic State-Space Models is popular models in infectious disease dynamics, which includes continuum models deterministic models (e.g., various types of differential equations), stochastic models (e.g., Markov-chain), complex networked models, and agent-based simulation. Empirical/Machine Learning-Based Models uses web-based data mining to identify epidemic trends.

Underreporting which severely impact the different epidemic models is prevalent, because of inadequate management and difficult environment for supervision. Wong et al. [2017] reviewed several models in terms of underreporting (or missing data). To handle missing data in the Ebola outbreaks, Bayesian frameworks that retain uncertainty could be used for a statistical outbreak formula to analyze the severity of infections.

In this research, we have reviewed literatures from the perspectives of a variety in analysis purposes (e.g., reproduction number estimation) for EVD as shown in Table 1. Such analyses have been implemented by various approaches (e.g., Bayesian Markov Chain Monte Carlo and Ordinary Differential Equations). Table 1 shows the list of the analysis approaches and analysis purposes.

**Table 1. Models for EVD dynamics**

| References | Analysis Approaches | Analysis Purposes |
|---|---|---|
| **[Polettoet al., 2014]** | GLobal Epidemic and Mobility (GLEaM) Model [Balcan et al., 2009] | International spread of the epidemic |
| **[Alizonet al., 2014]** | Bayesian Markov Chain Monte Carlo | Reproduction number |
| **[House, 2014]** | Bayesian Markov Chain Monte Carlo | Reproduction number & case fatality |
| **[Liu et al., 2015]** | Bayesian Markov Chain Monte Carlo | Reproduction number, turning point, & final size |
| **[Pruytet al., 2015]** | Bayesian Markov Chain Monte Carlo | Case fatality risk, hospitalization fatality risk, time-delay distributions, & prevalence of hospitalized cases |
| **[Shenet al., 2015]** | Statistics | Virus contact pattern |
| [Backer & Wallinga, 2016] | Bayesian Markov Chain Monte Carlo | Time-varying effective reproduction numbers over spatiotemporal resolution |
| [Maurer et al., 2014] | Bayesian Markov Chain Monte Carlo | Time-dependent district-specific effective reproduction number |
| [Pandey et al., 2014] | Continuous-time stochastic compartment model | Reproduction number |
| [Krauer et al., 2016] | Generalised linear mixed effects model | Reproduction number & growth rates |
| [Getz et al., 2015] | Markov transmission chain model | Reproduction number |
| [Lau et al., 2017] | Bayesian Markov Chain Monte Carlo | Age-specific spatiotemporal prediction |
| [Area et al., 2015] | Fractional calculus | SEIR (susceptible, exposed, infected, and recovered) |
| [Abbate et al., 2016] | Ordinary differential equations | SEICR (susceptible, exposed, infectious, convalescent, and recovered) |
| [Guo et al., 2016] | Ordinary differential equations | SIR (susceptible, infective, and removed) |
| [van den Driessche, 2017] | Ordinary differential equations | SEIR (susceptible, exposed, infected, and recovered) |
| [Evans & Mammadov, 2014] | Linear model | Reproduction numbers & time intervals |
| [Webb & Browne, 2016] | Probability model & ordinary differential equations | SIR (susceptible, infective, and removed) & disease age |
| [Berge et al., 2017] | Ordinary differential equations | SIR (susceptible, infective, and removed) & indirect transmissions |
| [Agusto et al., 2015] | Deterministic time-varying model | SIR (susceptible, infective, and removed) |
| [Nishiura & Chowell, 2015] | Probability model | Reproduction number, per-contact probability of infection, and growth rate |
| [Chowell & Nishiura, 2014] | Ordinary differential equations | Case fatality rate & impact of control interventions |
| [Barbarossa et al., 2015] | Deterministic time-varying model | SIHDEBR (susceptible, infectious, hospitalized, died, latent, buried, and removed) |
| [Rizzo et al., 2016] | Activity driven networks [Perra et al., 2012] | SEIHFRrRd (susceptible, exposed, infected, hospitalized, dead, recovered, and buried) |
| [Ngwa & Teboh-Ewungkem, 2016] | Ordinary differential equations | Human population into 11 states representing disease status |
| [Browne et al., 2014] | Ordinary differential equations | SEICiR (susceptible, incubating, infectious, contaminated, isolated, and removed) |
| [Abbate et al., 2016] | Ordinary differential equations | SIRD (susceptible, infected, removed, and deceased) with controls |
| [Al Darabsah & Yuan, 2016] | Ordinary differential equations | SEIRD (susceptible, exposed, infectious, recovered, and dead) |
| [Rachah & Torres, 2016] | Ordinary differential equations | SEIR (susceptible, exposed, infective, recovered) with controls |
| [Koya & Mamo, 2015] | Ordinary differential equations | SEIIhR (susceptible, exposed, infected, isolated, and removed) with controls |
| [Area et al., 2017] | Ordinary differential equations | SEIHRDBC (susceptible, exposed, infected, hospitalized, |
| | | asymptomatic but still infectious, dead but not buried, buried, and completely recovered) with controls |
| **[Althaus et al., 2015]** | Ordinary differential equations | SEIRD (susceptible, exposed, infectious, recover, and die) |
| **[Scarpino et al., 2014]** | Ordinary differential equations | SEIR (susceptible, exposed, infected, and recovered) |

The analysis purpose of most papers was to estimate the reproduction number and to build the compartment model (e.g., SEIR (susceptible, exposed, infected, and recovered)). Most analyses in the EVD domain are related to feature selection in machine learning. Finding related features, attributes, or factors as predictors is the subject of feature selection, but it is cumbersome to perform. From a feature selection perspective, EVD researchers have studied several topics such as spatiotemporal patterns, hospital case patterns, age-specific patterns, virus contact patterns, and international spread patterns.

To perform such analyses, various types of approaches were used. Basically, many variations of Ordinary Differential Equations were applied to estimate reproduction rates in a compartment model. To improve the prediction accuracy, incorporate various factors, and address missing data, novel approaches using Bayesian Markov Chain Monte Carlo (MCMC) were applied. Some special types of models (e.g., GLobal Epidemic and Mobility (GLEaM) [Balcan et al., 2009], and Activity Driven Networks (ADN) [Perra et al., 2012]) were introduced to represent complex structures in the real world and reason about outputs from such complex structures.

## 3. Collective Intelligence Multi-Model Integration Platform: Bayes Cloud

Understanding situations, making decisions, and addressing problems are obviously important tasks in our real life. For these tasks, modeling can be in the initial step and can play the important role. As many quoted words, "All models are wrong but some are useful [Box, 1979]", models can be useful to understand ideas of interest. In other words, the models are informative, although they are not exactly same as reality.

Practically, an existing model is useful to model developers on the assumption that the model does not mislead the reality. The roles of such existing model include (1) *illuminating*, (2) *sharing*, (3) *summarizing*, (4) *evaluating*, (5) *comparing*, (6) *inferring*, (7) *improving*, and (8) *creating*.

(1) **Illuminating:** The model helps us to quickly understand ideas, even if we do not have any prior knowledge.
(2) **Sharing:** The model which is explicitly described in a certain form helps us to efficiently share ideas.
(3) **Summarizing:** The model compresses information of the large and complex reality into a summary.
(4) **Evaluating:** The model, a clearly described idea, can be evaluated, but implicit idea cannot.
(5) **Comparing:** The models developed from multiple independent people can be explicitly compared.
(6) **Inferring:** The model enables us to reason about a query or question we have.
(7) **Improving:** The model helps us to improve our previous idea.
(8) **Creating:** The models can be the basis for creating new ideas.

As Einstein's statement ("Everything should be made as simple as possible, but not simpler"), a model should be simple, but not simpler. Our research focuses on the latter view. To address the real world, a model should be comprehensive, but not isolated. However, sometimes such comprehensiveness generates controversy by researchers and practitioners. Therefore, several different models (or hypotheses) should be introduced and compared. Our platform is developed to search different models, perform model integration, and allow model comparisons. Thereby, we try to increase the possibility for building the right model.

The platform performs two main functions: (1) *Model Development* and (2) *Model Application*. In the model development, BN models are created and edited by SME and/or AI crawlers, learned by Machine Learning, and integrated by SME and/or a model integration algorithm. In the model application, BN models are used to show the phenomena to model users, share them among model users, and reason about the queries from model users.

### 3.1. Main Functions of Bayes Cloud

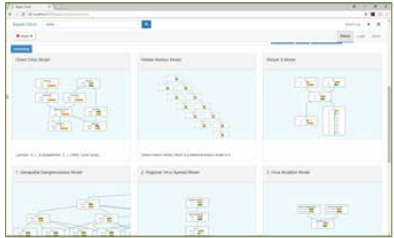
Searching

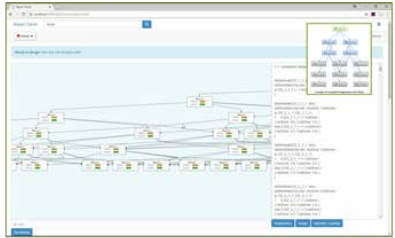
Editing

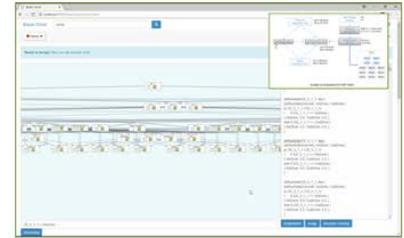
Integration

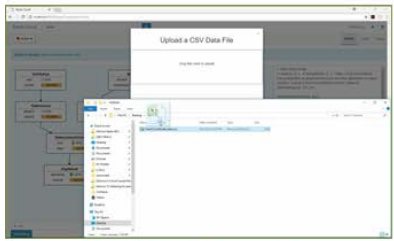
Machine Learning

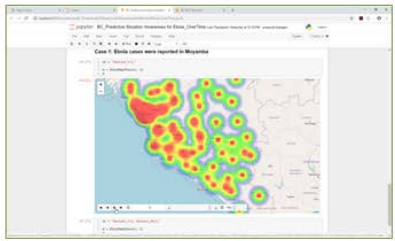
Reasoning

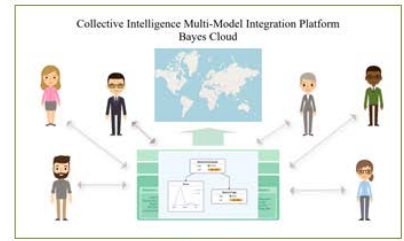
Sharing

#### 3.1.1. Model Editing

Model editing allows users to build a BN model or modify existing BN models using a Bayesian network script code. The Bayesian network script [Sun et al., 2011] is an intuitive and user-friendly language to let users easily write codes to design BN models and apply machine learning. Bayes Cloud provides a model editing module in which the Bayesian network script can be written to define a node containing probability distribution and causal relationships between nodes. For example, in the demo, the node A is first defined in Line 1. The type of states in the node A is defined as discrete and its two states (a1 and a2) are specified in Line 3. Probability values for these states are specified as 0.2 and 0.8 respectively in Line 5, p. Users can re-use existing nodes to create a new node with similar structure and modify as needed. From Line 8 to Line 16, the node B is defined by copying the node A. The difference in the node B is that the conditional probability distributions are specified by using if-statement to create a relationship to node A. The probability of b1 given a1 as A is 0.2 while the probability of b2 given a1 as A is 0.8. The probability of b1 given a2 as A is 0.8 while the probability of b2 given a2 as A is 0.2. Similarly, users can easily create a simple to complex BN model.

#### 3.1.2. Model Integration

Without loss of generality, model integration can be viewed as a process for merging two BNs ("$BN_1$" and "$BN_2$") in order to obtain a third model $BN_{merged}$. The resulting model shall embody the information from the original models, and also support joint reasoning unattainable solely by the individual models.

This integration process can be performed either manually or automatically. Simply stated, in the manual approach the SME loads and edits the two BNs in the Bayes Cloud workspace, and $BN_{merged}$ is obtained by reusing all random variables (a.k.a. nodes) and including new relationships (a.k.a. arcs), with respective edits to

4**11**conditional probability distributions. In contrast, statistic/probabilistic models and algorithms are applied in the automatic approach in order to generate the $BN_{merged}$ with small or no SME intervention. Generally, the model resulting from the automatic approach can be seen as a statistical approximation to both $BN_1$ and $BN_2$.

The automatic model integration function shall assume that the pair $BN_1$ and $BN_2$ share common random variables $V_{shared}$. If the BNs don't share any common random variables (*i.e.*, when $V_{shared}$ is an empty set), then all the variables in $BN_1$ will be considered as independent from the variables in $BN_2$ and vice-versa. In such situation, the $BN_{merged}$ basically comprises two disconnected subnets, with each subnet being $BN_1$ and $BN_2$. Although this case is not technically different from keeping the two original BNs separate, we'll keep this case (the model comprising disconnected subnets $BN_1$ and $BN_2$) as a valid model in the repository, as a means for explicitly indicating that variables in $BN_1$ are independent from variables in $BN_2$[*].

Non-linear optimization methods can be applied in order to automatically find the joint probability distribution of $BN_{merged}$ from the probability distributions of $BN_1$ and $BN_2$, when there are common variables (*i.e.*, when $V_{shared}$ is a non-empty set). Simply stated, this is performed by using non-linear optimzer tools under the following criteria:

I. Represent probabilities of joint states (product space of the variables in $BN_1$ and $BN_2$ together) as optimization variables to be considered in the non-linear optimization. Since these joint state variables are probabilities, we shall also include constraints to guarantee that these joint state variables are non-negative and adds up to 1.

II. In terms of the above joint state variables, model an objective function (function to be optimized by the non-linear optimizer) based on information entropy criteria, namely tominimize KL-divergence [Kullback& Leibler, 1951], in order to find a joint probability which best approximates (in terms of information entropy) with the probability distributions of $BN_1$ and $BN_2$.

III. After obtaining the results from the optimizer, use the new joint probability to build back the structure of $BN_{merged}$. The set of variables in $BN_{merged}$ is the union of the variables in $BN_1$ and $BN_2$, so this step reduces to reconstruction of the relationships (a.k.a. arcs)[†].

The above non-linear optimization solution may become infeasible, as it works with joint states—a product space of the random variables. Even though such joint states can be factorized by exploiting conditional (in)dependence [Jensen et al., 1994], and thus the scope of the non-linear optimization can be reduced to such factored structure, there will be complex cases which are infeasible even with such factorization. In addition, the above method is unapplicable to continuous, uncountable, real number variables (because in such situation the number of joint states will become uncountable/infinite). Under such situation, Bayes Cloud shall use simulation-based approaches with following adaptations:

I. Randomly pick one BN (either $BN_1$ or $BN_2$) to start the sampling (a.k.a. simulation).

II. Sample (*i.e.*, generate a plausible value) all the variables in the picked BN by using any sampling method (e.g., Gibbs sampling Bayesian inference algorithm [Gelfand et al., 1990]). Please, note that this shall also sample variables in $V_{shared}$.

---

[*] When $BN_1$ and $BN_2$ were managed separately, the independence property was unknown/unstated. On the other hand, in $BN_{merged}$ the independence is made explicit.

[†] A formal way of re-building the arcs of $BN_{merged}$ from its joint probability distribution is to test the conditional (in)dependenceof the variables in order to decide the presence/absence of arcs and its direction. A simpler, but reasonable way is to consider that arcs in $BN_{merged}$ are also a union of the arcs in $BN_1$ and $BN_2$.

III. Similarly, perform sampling of variables in the other BN (the one not picked in the first step), but conditioned to values in $V_{shared}$ which were sampled in previous step.

IV. Repeat the above steps until obtaining a reasonable amount of sample sets.

V. Build $BN_{merged}$ by applying a Bayesian learning algorithm with the above sample sets as input data.

Besides of exhibiting an overhead due to recurrence to a Bayesian learning algorithm at the end of the process, the above simulation approach will result in $BN_{merged}$ which approximates $BN_1$ and $BN_2$, with a precision proportional to the number of sample sets (a.k.a. number of simulations). Although these characteristics can potentially cause large imprecision in the resulting model, due to accumulated imprecision/errors from simulation and re-learning, this general approach can be applied even in hybrid models (mixture of discrete and continuous random variable models) as well.

### 3.1.3. Causal Model Learning from Texts

The AI crawler collects information from research papers and uses that text information to extrapolate the causal information from it to create BN models.

### 3.1.4. Parameter Learning from Data

Given a BN structure and a set of training data, the platform performs Bayesian network parameter learning. The current version allows the continuous and discrete parameter learning.

### 3.1.5. Structure Learning from Data

Given a set of training data, the platform performs Bayesian network structure learning. In other words, correlations between random variables are found by fitting the training data. After this, Bayesian network parameter learning is performed automatically to find parameters of the random variables.

### 3.1.6. Model Reasoning

Model reasoning is an application in the modeling platform which uses various reasoning information from new evidence to change the models in real-time and as needed. Users also use a simple script to specify a node name and its state based on evidence in the bottom panel. Assuming the state of the node A is a1 based on evidence, users would write a code A = a1. Then remaining is quite simple. Users just need to click the Reasoning button and will see the instant result from the BN model. As users obtain a new evidence, they will simply change the code to update A = a2 and click the Reasoning button to update the result. In case of supervised reasoning, users should be able to measure reliability and accuracy of model in a real time basis. In case of complex BN models, users can assign states to multiple nodes in a similar manner. As states change based on evidence, users can easily modify the script and get an instant result by using model reasoning functionality in the modeling platform.

### 3.2. Main Functions of Cloud Service

### 3.2.1. Model Searching

Registered Models are subject to search by keywords in the modeling platform as well as various search engines in the market. Model searching allows the user to find existing models in the Bayesian cloud with a keyword that is located at the top. This can be used with or without other models. Searched models can be loaded into the modeling platform for editing, reasoning, integrating, and machine learning. Users will be able to find the model using other web search engines such as Google outside of the modeling platform. Even though users outside of the modeling platform might not be able to perform functions in the modeling platform, they will be able to see details about the model as a web document that the original creator produced.

### 3.2.2. Model Sharing

When the BN model is built and registered in the modeling platform, the model can be searched and shared by other users using various platforms such as Python and R as well as the modeling platform. Modeling sharing is the most powerful function in the modeling platform because shared models can be further developed into more complex and reliable models by editing and integrating with individual or collective efforts. While the original producer of a shared model shall be credited by those who are willing to use the model, the number of BN models will grow exponentially as more users contribute to developing BN models in the modeling platform. The Bayesian network script code used in BN model will be compiled in other platforms as well. For example, a Bayes package will be developed for various development languages such as Python and R. Python developers will be able to install the Bayes package that will allow Python programmers to execute the Bayesian network script code chunks in various IDEs such as Jupyter Notebook. Similarly, R programmers will be able to install the Bayes package and run the code chunks in R Notebook.

### *3.2.3. Model Information Registering & Editing*

Model registering allows user to register their own BN models in Bayes Cloud. Registered models can be used later by other users as they can edit the model or integrate into other models. The registering process is quite simple. After a BN model is built, users can start registering by clicking the registration button. And then a registration panel appears and the users enter details for the model such as title, description, author, and search keywords. When the model is registered by clicking a register button in the registration panel, the model information as well as the model image will be stored in the Bayes Cloud database. The users can retrieve the model from the database for further editing and reasoning and delete their models.

## 4. Use Case for the Ebola Disease BN Model

In this section, we introduce one illustrative use case of Bayes Cloud for the Ebola Virus Disease (EVD). We introduce a model that integrates independent models for the EVD spread to show how to use Bayes Cloud in the infectious disease domain. To understand the Ebola epidemic, holistic situational awareness is required. To do this, we need to integrate multiple sources of data such as research papers, medical datasets, and expert knowledge. Figure 3 shows an illustrative example of multiple sources regarding EVD.

EVD research papers may contain summary statistics, causal relationships among epidemic factors, estimated parameters for factors, structured models, and links for data sources. Using this information, a specific-purpose Bayesian network (BN) in Section 2 can be developed. The specific-purpose BN means the BN which is built to answer a specific question. The examples for the specific-purpose BN include a BN for a compartment model (e.g., SEIR (susceptible, exposed, infected, and recovered)), a BN for predicting a serial interval (i.e., the time between the primary case and the secondary case for EVD onset), a BN for a transmission model (e.g., between reservoirs and human), a BN for representing the indirect transmission, a BN for identifying the biological characteristics between the pathogen and the host, a BN for describing the detailed contact patterns, and a BN for global epidemic spread (including traveller patterns). From these specific-purpose BNs, one integrated model can be constructed. The integrated model provides the holistic situational awareness for the Ebola epidemic and would be the basis of decision making for EVD to prevent further disease spread and minimize any risk of getting infected.

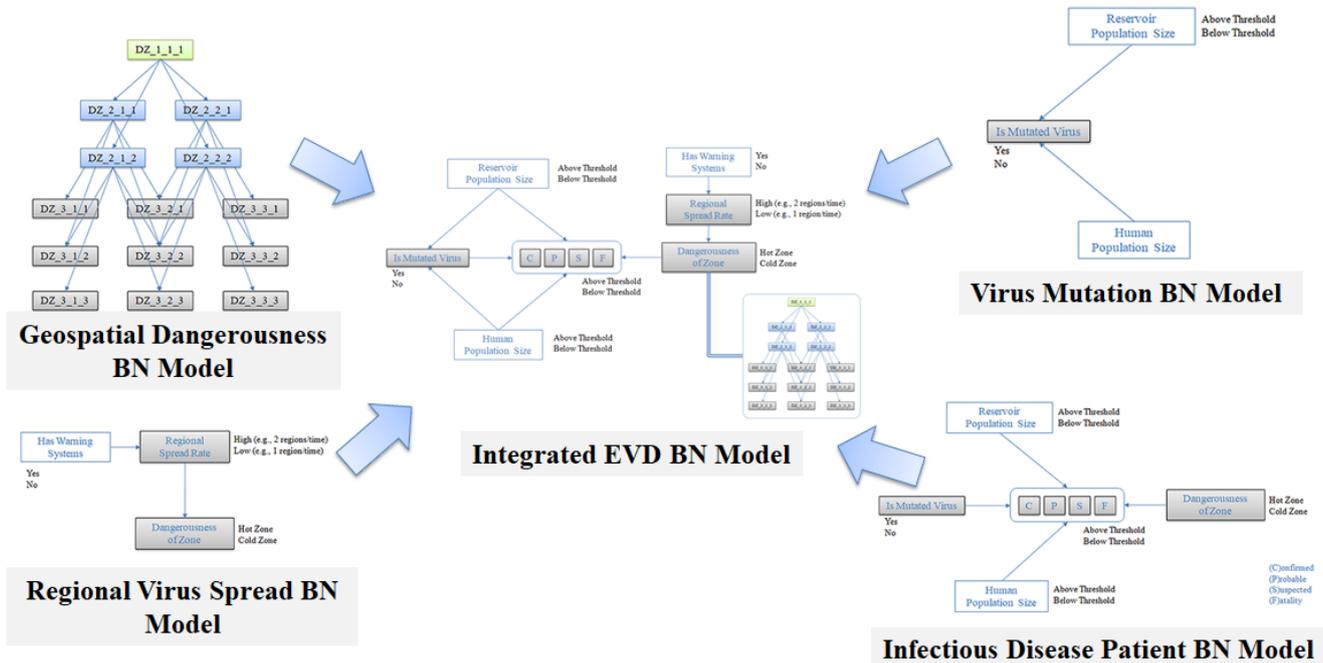

Figure 3. Concept of BN Model Integration for EVD (This figure will be changed.)

In this research, the main specific-purpose BNs for EVD include (1) *Geospatial Dangerousness Model*, (2) *Regional Virus Spread Model*, (3) *Virus Mutation Model*, and (4) *Infectious Disease Patient Model*. These are derived from several literatures on the subject. We also introduce an integrated model constructed from these models.

*4.1. Geospatial Dangerousness BN Model*

The geospatial dangerousness BN model represents properties of sub-regions in a certain region and relationships between nearby sub-regions. Geospatial models can be used to predict the EVD spread. For the EVD spread, many geospatial models were proposed in the literature [Balcan et al., 2009][Carroll et al., 2015][Kiskowski & Chowell, 2016][Lau et al., 2017][Fiorillo et al., 2018]. In our research, we introduce a special BN structure to form a geospatial map. It seems like a pyramid structure, in which a root node is associated with its child nodes and the child nodes are associated with their sub-child nodes, and so forth. The advantage of this model is that the relationship between the sub-regions can be flexibly expressed and the relationship between an upper-level region and lower-level regions can be seamlessly represented. By doing so, the model can be easily used to represent and reason about the spread of an epidemic disease.

The following figure is an example of the geospatial dangerousness BN which separates and categorizes the regions and sub-regions. In each node has a designated name for the specific region (e.g., New York and Virginia) and is associated with the region property. For example, the infection level can be such region property and may have a discrete value (e.g., hot or cold zone) or a continuous value (e.g., {$r$ as risk | $r >= 0$ and $r$ is real}). The word DZ in the node name stands for *Dangerousness of Zone*. The numbers following it indicate the depth, x location, and y location of the region, respectively. Some reports or findings pertaining to the infection is inputted into the model showing which then predicts those regions as hot or cold zones depending on the level of infection. For example, if the region DZ_3_1_3 has been reported as an infectious case, the region is set as hot zone and its updating influences other zones (e.g., DZ_3_1_2, DZ_3_2_2, DZ_3_2_3, and so forth).

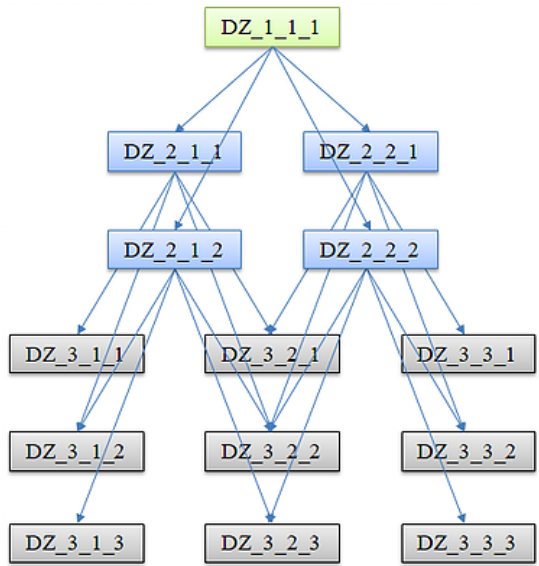

**Figure 4. Example of Geospatial Dangerousness BN Model**

Lau et al. [2017] estimated the EVD transmission over regions for an infector as median value of 2.51 km. In other words, when one infector happens, the influence of the infector spreads to the median range. Such spread rate should be applied to the geospatial dangerousness BN, especially the conditional probability distribution in a node. The conditional probability distribution for each node can be simply assigned with a *k* value, where $1 > k > 0.5$. Such *k* value can be estimated using the spread rate. The following distribution shows an illustrative example, where the child region has the probability of hot zone = *k*, if a patient region is hot zone.

$$P(DZ_{child} = hot\ zone | DZ_{parent} = hot\ zone) = \{hot\ zone: k; cold\ zone: 1 - k; \}.$$

Basically, the geospatial dangerousness BN is developed without any control. Therefore, the same degree of influence is applied to any node in the same depth (or level). However, obviously, regional characteristics and other control can influence the spread rate. In the following sub-section, we discuss about this.

*4.2. Regional Virus Spread BN Model*

The virus spread rate is affected by control (e.g., alarm broadcasting and environment characteristics). Althaus [2014] researched the spread of an infection for EVD in the absence and the presence of control interventions. Public health intervention can significantly reduce the epidemic spread. Such intervention, control, and environment characteristic can be included in a BN model. We define a simple BN model called Regional Virus Spread Model, which represents the factors affecting the spread rate. For example, a region of high GDP (Gross Domestic Product) may have an abundance of healthcare infrastructures. Such regional characterises, then, might prevent the rapid infectious disease spread.

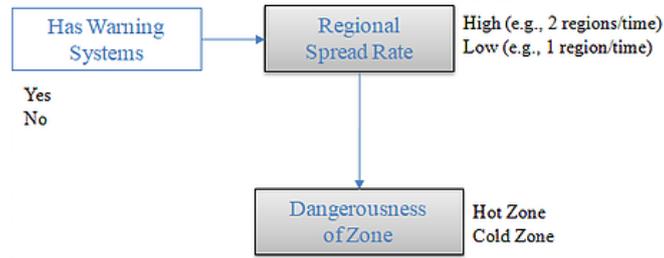

**Figure 5. Example of Regional Virus Spread BN Model**

The figure above shows the regional virus spread BN model representing the spread rate of EVD in a certain region and the factors involved in influencing it. The non-highlighted node *Has Warning Systems* indicates a known report that can be provided on site and influences the highlighted node *Regional Spread Rate*. Depending on if a region has access to the equipment needed to survey an area for biological hazards, this would be a warning system that would help in the prevention of spreading a virus. The regional spread rate indicates how fast the virus will spread being either high or low. The node *Dangerousness of Zone*, which has two states hot and cold zones, indicates the degree of risk at a region that is high or low in the amount of infected individuals by the virus.

*4.3. Virus Mutation BN Model*

Some virus species are fatal to humans. In 2014, especially three Ebola virus species (EBOV, Sudan Ebola virus, and Bundibugyo Ebola virus) were main viruses causing the high fatality rate of 30% to 90% [Baize et al., 2014]. Ebola virus species were mutated throughout the transmission between reservoirs. Potential reservoirs of EBOV were fruit bats which existed in most parts of West Africa. This is one of the reasons EBOV had occurred in the region. The existence of EBOV reservoirs causes the outbreak as well as the number of reservoirs affects the mutation rate. For some Ebola virus species, humans were reservoirs, and for other species, wildlife was reservoirs.

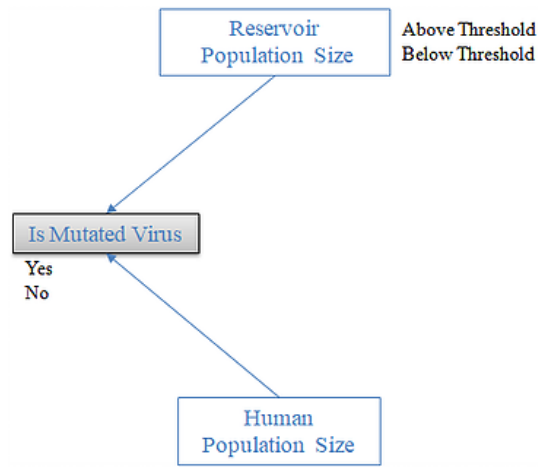

**Figure 6. Example of Virus Mutation BN Model**

The figure above is a simplified model to represent the chance of mutation in the Ebola virus and the factors involved in influencing it. The non-highlighted nodes indicate known reports that can be provided on site, and influence the highlighted node *Is Mutated Virus*. The basic principle of the model is that the higher the population of either animals or humans the higher the chance for the Ebola virus to mutate from the original virus.

## 4.4. Infectious Disease Patient BN Model

Age is an important demographic forecasting variable for the spread of EVD [Lau et al., 2017]. For example, the age groups below 15 and older than 45 have higher instantaneous hazard. And it was shown that the higher instantaneous hazard group with a long infectious period was estimated as a key driver of the EVD spread. Also, social contact and personal conditions could be important predictors. We define an infectious disease patient BN model, which represents a compartment model (e.g., SEIR (susceptible, exposed, infected, and recovered)).

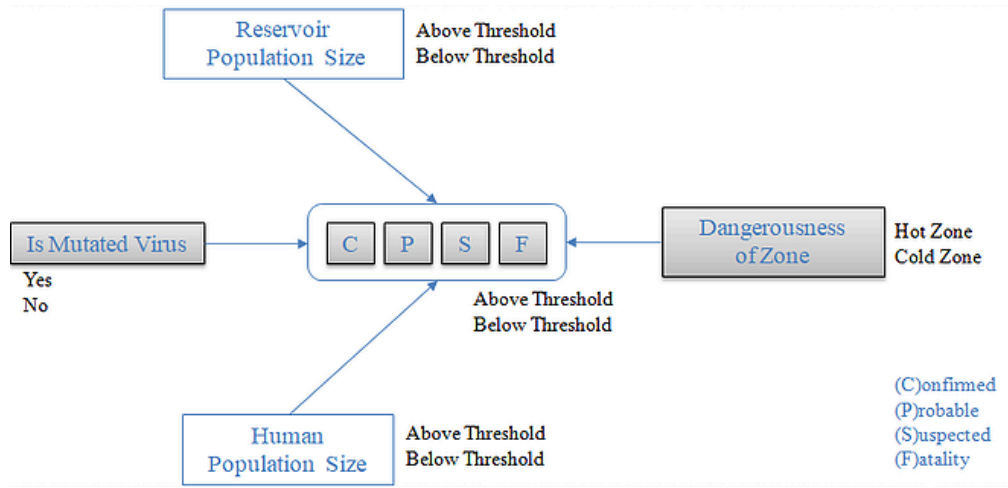

**Figure 7. Example of an Infectious Disease Patient BN Model**

Figure 7 shows an example of the infectious disease patient BN model. This model represents the number of patients categorized into (C)onfirmed, (P)robable, (S)uspected, and (F)atality individuals infected by the Ebola virus. Reservoir population size is referring to the number of local animal reservoir (e.g., bat) that may be the origin of Ebola virus [Morse et al., 2012]. Human population impacts the nodes C/P/S/F. The virus type impacts the nodes C/P/S/F as well, because if the virus is highly contagious, it will influence the number of probable and confirmed individuals, or if the virus has highly fatal symptoms, it will raise the number of fatalities.

## 4.5. Integrated EVD BN Model

Through sub-sections 4.1~4, four BN models for EVD were introduced. In this sub-section, we introduce the integrated BN model from these four BN models (the geospatial dangerousness BN model, regional virus spread BN model, virus mutation BN model, and infectious disease patient BN model).

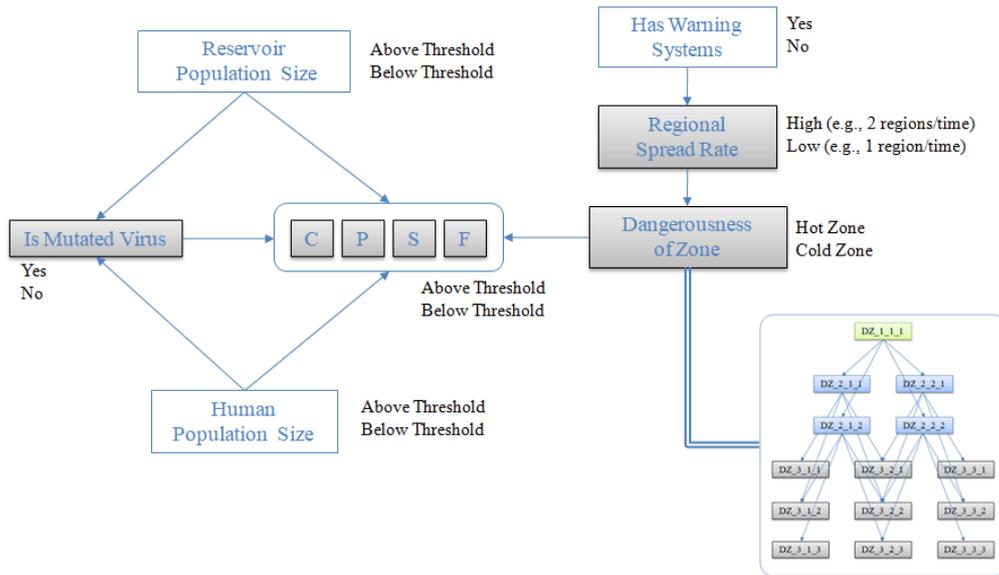

**Figure 8. Example of an Integrated EVD BN Model**

Figure 8 shows the integrated EVD BN model, which involves and combines all the other models. The major impact that differentiates this integrated model from the previous models is that now we see how a single region can be associated with other regions through the geospatial dangerousness BN model. Also, the human and animal populations simultaneously impact the virus mutation and the nodes C/P/S/F. And how the regional spread rate is applied to the each region.

The main benefit of the integrated model is holistic situational awareness in which we can now view how each and every node is affecting one another and/or all at once. The integration model is available in the following operating scenarios.

**Operational scenario:**

1. Development of the integrated EVD BN model

    a. For a certain region containing several sub-regions, the integrated EVD BN model is developed.

    b. Each specific-purpose BN is learned from collecting data and developed by experts and model/parameters from literatures.

    c. Specific-purpose BNs are integrated into the integrated EVD BN model.

    d. The integrated EVD BN model is applied to a disease surveillance system.

2. Situation awareness using the integrated EVD BN model

    a. Regional information (e.g., reservoir/human population size and warning systems) is inputted to the integrated EVD BN model.

    b. The numbers of C/P/S/F reported from a certain region are inputted to the integrated EVD BN model.

    c. The integrated EVD BN model predicts the risk level of surrounding regions of the reported region.

    d. A decision maker decides a proper treatment from several options (e.g., evacuation, vaccination, and containment) using predicted results.

# 5. Conclusions

The main benefit of the integrated model is holistic situational awareness in which we can now view how each and every node is affecting one another and/or all at once. The integration model is available in the following operating scenarios.

**Operational scenario:**

1. Development of the integrated EVD BN model

   a. For a certain region containing several sub-regions, the integrated EVD BN model is developed.

   b. Each specific-purpose BN is learned from collecting data and developed by experts and model/parameters from literatures.

   c. Specific-purpose BNs are integrated into the integrated EVD BN model.

   d. The integrated EVD BN model is applied to a disease surveillance system.

2. Situation awareness using the integrated EVD BN model

   a. Regional information (e.g., reservoir/human population size and warning systems) is inputted to the integrated EVD BN model.

   b. The numbers of C/P/S/F reported from a certain region are inputted to the integrated EVD BN model.

   c. The integrated EVD BN model predicts the risk level of surrounding regions of the reported region.

   d. A decision maker decides a proper treatment from several options (e.g., evacuation, vaccination, and containment) using predicted results.

REFERENCES

Abbate, J. L., Murall, C. L., Richner, H., & Althaus, C. L. (2016). Potential impact of sexual transmission on Ebola virus epidemiology: Sierra Leone as a case study. PLoS neglected tropical diseases, 10(5), e0004676.

Agusto, F. B., Teboh-Ewungkem, M. I., & Gumel, A. B. (2015). Mathematical assessment of the effect of traditional beliefs and customs on the transmission dynamics of the 2014 Ebola outbreaks. BMC medicine, 13(1), 96.

Ajelli, M., Parlamento, S., Bome, D., Kebbi, A., Atzori, A., Frasson, C., ... & Merler, S. (2015). The 2014 Ebola virus disease outbreak in Pujehun, Sierra Leone: epidemiology and impact of interventions. BMC medicine, 13(1), 281.

Al-Darabsah, I., & Yuan, Y. (2016). A time-delayed epidemic model for Ebola disease transmission. Applied Mathematics and Computation, 290, 307-325.

Alizon, S., Lion, S., Murall, C. L., & Abbate, J. L. (2014). Quantifying the epidemic spread of Ebola virus (EBOV) in Sierra Leone using phylodynamics. Virulence, 5(8), 825-827.

Althaus, C. (2015). Ebola superspreading. Lancet infectious diseases, 15(5), 507-508.

Althaus, C. L. (2014). Estimating the reproduction number of Ebola virus (EBOV) during the 2014 outbreak in West Africa. PLoS currents, 6.

Althaus, C. L. (2015). Rapid drop in the reproduction number during the Ebola outbreak in the Democratic Republic of Congo. PeerJ, 3, e1418.

Althaus, C. L., Low, N., Musa, E. O., Shuaib, F., & Gsteiger, S. (2015). Ebola virus disease outbreak in Nigeria: transmission dynamics and rapid control. Epidemics, 11, 80-84.

Andreassen, S., Woldbye, M., Falck, B., & Andersen, S. K. (1987, August). MUNIN: A causal probabilistic network for interpretation of electromyographic findings. In Proceedings of the 10th international joint conference on Artificial intelligence-Volume 1 (pp. 366-372). Morgan Kaufmann Publishers Inc..

Area, I., Batarfi, H., Losada, J., Nieto, J. J., Shammakh, W., & Torres, Ã□. (2015). On a fractional order Ebola epidemic model. Advances in Difference Equations, 2015(1), 278.

Area, I., NdaÃ□rou, F., Nieto, J. J., Silva, C. J., & Torres, D. F. (2017). Ebola model and optimal control with vaccination constraints. arXiv preprint arXiv:1703.01368.

Backer, J. A., & Wallinga, J. (2016). Spatiotemporal analysis of the 2014 Ebola epidemic in West Africa. PLoS computational biology, 12(12), e1005210.

Baize, S., Pannetier, D., Oestereich, L., Rieger, T., Koivogui, L., Magassouba, N. F., ... & Tiffany, A. (2014). Emergence of Zaire Ebola virus disease in Guinea. New England Journal of Medicine, 371(15), 1418-1425.

Balcan, D., Colizza, V., Gonçalves, B., Hu, H., Ramasco, J. J., & Vespignani, A. (2009). Multiscale mobility networks and the spatial spreading of infectious diseases. Proceedings of the National Academy of Sciences, pnas-0906910106.

Barbarossa, M. V., DÃ©nes, A., Kiss, G., Nakata, Y., RÃ¶st, G., & Vizi, Z. (2015). Transmission dynamics and final epidemic size of Ebola virus disease outbreaks with varying interventions. PloS one, 10(7), e0131398.

BCM. (2018). [Online]. Available at https://www.bcm.edu/departments/molecular-virology-and-microbiology/emerging-infections-and-biodefense/introduction-to-infectious-diseases

Berge, T., Lubuma, J. S., Moremedi, G. M., Morris, N., & Kondera-Shava, R. (2017). A simple mathematical model for Ebola in Africa. Journal of biological dynamics, 11(1), 42-74.

Browne, C., Gulbudak, H., & Webb, G. (2015). Modeling contact tracing in outbreaks with application to Ebola. Journal of theoretical biology, 384, 33-49.

Browne, C., Huo, X., Magal, P., Seydi, M., Seydi, O., & Webb, G. (2014). A model of the 2014 ebola epidemic in west africa with contact tracing. arXiv preprint arXiv:1410.3817.

Browne, C., Huo, X., Magal, P., Seydi, M., Seydi, O., & Webb, G. (2014). A model of the 2014 ebola epidemic in west africa with contact tracing. arXiv preprint arXiv:1410.3817.

Box, G. E. (1979). Robustness in the strategy of scientific model building. In Robustness in statistics (pp. 201-236).

Camacho, A., Kucharski, A., Aki-Sawyerr, Y., White, M. A., Flasche, S., Baguelin, M., ... & Tiffany, A. (2015). Temporal changes in Ebola transmission in Sierra Leone and implications for control requirements: a real-time modelling study. PLoS currents, 7.

Carroll, M. W., Matthews, D. A., Hiscox, J. A., Elmore, M. J., Pollakis, G., Rambaut, A., ... & Abdellati, S. (2015). Temporal and spatial analysis of the 2014–2015 Ebola virus outbreak in West Africa. Nature, 524(7563), 97.

CDC. (2018). [Online]. Available at https://www.cdc.gov/vhf/ebola/history/2014-2016-outbreak/case-counts.html.

Chowell, G., & Nishiura, H. (2014). Transmission dynamics and control of Ebola virus disease (EVD): a review. BMC medicine, 12(1), 196.

Chowell, G., & Nishiura, H. (2015). Characterizing the transmission dynamics and control of ebola virus disease. PLoS biology, 13(1), e1002057.

Chowell, G., Sattenspiel, L., Bansal, S., & Viboud, C. (2016). Mathematical models to characterize early epidemic growth: A review. Physics of life reviews, 18, 66-97.

Chowell, G., Viboud, C., Hyman, J. M., & Simonsen, L. (2015). The Western Africa ebola virus disease epidemic exhibits both global exponential and local polynomial growth rates. PLoS currents, 7.

Chowell, G., Viboud, C., Simonsen, L., & Moghadas, S. M. (2016). Characterizing the reproduction number of epidemics with early subexponential growth dynamics. Journal of The Royal Society Interface, 13(123), 20160659.

Chowell, G., Viboud, C., Simonsen, L., Merler, S., & Vespignani, A. (2017). Perspectives on model forecasts of the 2014â€'2015 Ebola epidemic in West Africa: lessons and the way forward. BMC medicine, 15(1), 42.

Chretien, J. P., Riley, S., & George, D. B. (2015). Mathematical modeling of the West Africa Ebola epidemic. Elife, 4, e09186.

Laskey, K. B. (2008). MEBN: A Language for First-Order Bayesian Knowledge Bases. Artificial Intelligence, 172(2-3).

Pearl, J., & Mackenzie, D. (2018). The Book of Why: The New Science of Cause and Effect. Basic Books.

Cleaton, J. M., Viboud, C., Simonsen, L., Hurtado, A. M., & Chowell, G. (2015). Characterizing Ebola transmission patterns based on internet news reports. Clinical Infectious Diseases, 62(1), 24-31.

Conrad, J. R., Xue, L., Dewar, J., & Hyman, J. M. (2016). Modeling the Impact of Behavior Change on the Spread of Ebola. In Mathematical and Statistical Modeling for Emerging and Re-emerging Infectious Diseases (pp. 5-23). Springer, Cham.

Cooper, G. F. (1984). *NESTOR: A Computer-Based Medical Diagnostic Aid That Integrates Causal and Probabilistic Knowledge* (No. STAN-CS-84-1031). STANFORD UNIV CA DEPT OF COMPUTER SCIENCE.

Curran, K. G. (2016). Cluster of Ebola virus disease linked to a single funeralâ€'Moyamba District, Sierra Leone, 2014. MMWR. Morbidity and mortality weekly report, 65.

Diez, F. J., Mira, J., Iturralde, E., & Zubillaga, S. (1997). DIAVAL, a Bayesian expert system for echocardiography. Artificial Intelligence in Medicine, 10(1), 59-73.

DJIOMBA NJANKOU, S. D., & Nyabadza, F. (2016). An optimal control model for Ebola virus disease. Journal of Biological Systems, 24(01), 29-49.

Evans, R. J., & Mammadov, M. (2014). Dynamics of Ebola epidemics in West Africa 2014. F1000Research, 3.

Fasina, F. O., Shittu, A., Lazarus, D., Tomori, O., Simonsen, L., Viboud, C., & Chowell, G. (2014). Transmission dynamics and control of Ebola virus disease outbreak in Nigeria, July to September 2014. Eurosurveillance, 19(40), 20920.

Faye, O., Boëlle, P. Y., Heleze, E., Faye, O., Loucoubar, C., Magassouba, N. F., ... & Cauchemez, S. (2015). Chains of transmission and control of Ebola virus disease in Conakry, Guinea, in 2014: an observational study. The Lancet Infectious Diseases, 15(3), 320-326.

Fenton, N., & Neil, M. (2010). Comparing risks of alternative medical diagnosis using Bayesian arguments. Journal of Biomedical Informatics, 43(4), 485-495.

Ferrara, E. (2017). Contagion dynamics of extremist propaganda in social networks. Information Sciences, 418, 1-12.

Fine, P. E. (2003). The interval between successive cases of an infectious disease. American journal of epidemiology, 158(11), 1039-1047.

Fiorillo, G., Bocchini, P., & Buceta, J. (2018). A Predictive Spatial Distribution Framework for Filovirus-Infected Bats. Scientific reports, 8(1), 7970.

Fisman, D., & Tuite, A. (2014). Projected impact of vaccination timing and dose availability on the course of the 2014 West African Ebola epidemic. PLoS currents, 6.

Gardy, J. L., & Loman, N. J. (2018). Towards a genomics-informed, real-time, global pathogen surveillance system. Nature Reviews Genetics, 19(1), 9.

Gelfand, A., Hills, S., Racine-Poon, A. and Smith, A. (1990). Illustration of Bayesian inference in normal data models using Gibbs sampling. J. Amer. Statist. Assoc. 85. 972–982

Getz, W. M., Gonzalez, J. P., Salter, R., Bangura, J., Carlson, C., Coomber, M., ... & Wauquier, N. (2015). Tactics and strategies for managing Ebola outbreaks and the salience of immunization. Computational and mathematical methods in medicine, 2015.

Gire, S. K., Goba, A., Andersen, K. G., Sealfon, R. S., Park, D. J., Kanneh, L., ... & Wohl, S. (2014). Genomic surveillance elucidates Ebola virus origin and transmission during the 2014 outbreak. science, 1259657.

Grassly, N. C., & Fraser, C. (2008). Mathematical models of infectious disease transmission. Nature Reviews Microbiology, 6(6), 477.

Grigorieva, E. V., & Khailov, E. N. (2015). Optimal intervention strategies for a SEIR control model of Ebola epidemics. Mathematics, 3(4), 961-983.

Guo, Z., Xiao, D., Li, D., Wang, X., Wang, Y., Yan, T., & Wang, Z. (2016). Predicting and evaluating the epidemic trend of ebola virus disease in the 2014-2015 outbreak and the effects of intervention measures. PloS one, 11(4), e0152438.

Guo, Z., Xiao, D., Li, D., Wang, X., Wang, Y., Yan, T., & Wang, Z. (2016). Predicting and evaluating the epidemic trend of ebola virus disease in the 2014-2015 outbreak and the effects of intervention measures. PloS one, 11(4), e0152438.

Gutfraind, A., & Meyers, L. A. (2015). Evaluating large-scale blood transfusion therapy for the current Ebola epidemic in Liberia. The Journal of infectious diseases, 211(8), 1262-1267.

Hall, M. D., Woolhouse, M. E. J., & Rambaut, A. (2016). Using genomics data to reconstruct transmission trees during disease outbreaks. Revue scientifique et technique (International Office of Epizootics), 35(1), 287.

Heckerman, D., Horvitz, E., & Nathwani, B. (1992). Toward Normative Expert Systems: Part I, the Pathfinder Project. Knowledge Systems Laboratory, Medical Computer Science.

Holmes, E. C., Dudas, G., Rambaut, A., & Andersen, K. G. (2016). The evolution of Ebola virus: Insights from the 2013‒2016 epidemic. Nature, 538(7624), 193.

Holmes, E. C., Dudas, G., Rambaut, A., & Andersen, K. G. (2016). The evolution of Ebola virus: Insights from the 2013–2016 epidemic. Nature, 538(7624), 193.

House, T. (2014). Epidemiological dynamics of Ebola outbreaks. Elife, 3, e03908.

Hsieh, Y. H. (2015). Temporal course of 2014 Ebola virus disease (EVD) outbreak in West Africa elucidated through morbidity and mortality data: a tale of three countries. PloS one, 10(11), e0140810.

J. Pearl, Probabilistic Reasoning in Intelligent Systems: Networks of Plausible Inference, Revised second printing. Morgan Kaufmann, 1988/2014.

Jensen, F. V., & Jensen, F. (1994). Optimal junction trees. In Uncertainty Proceedings 1994 (pp. 360-366).

K. P. Burnham and D. R. Anderson, "Information and Likelihood Theory: A Basis for Model Selection and Inference," in Model Selection and Multimodel Inference: A Practical Information-Theoretic Approach, New York, Springer, 2007, pp. 49-96.

Kaasik-Aaslav, K., & Coulombier, D. (2015). The tail of the epidemic and the challenge of tracing the very last Ebola case. Eurosurveillance, 20(12), 21075.

Kaner, J., & Schaack, S. (2016). Understanding Ebola: the 2014 epidemic. Globalization and health, 12(1), 53.

Keeling, M. J., & Danon, L. (2009). Mathematical modelling of infectious diseases. British Medical Bulletin, 92(1).

Khan, A., Naveed, M., Dur-e-Ahmad, M., & Imran, M. (2015). Estimating the basic reproductive ratio for the Ebola outbreak in Liberia and Sierra Leone. Infectious diseases of poverty, 4(1), 13.

Kiskowski, M. (2014). Description of the early growth dynamics of 2014 West Africa Ebola epidemic. arXiv preprint arXiv:1410.5409.

Kiskowski, M. A. (2014). A three-scale network model for the early growth dynamics of 2014 West Africa Ebola epidemic. PLoS currents, 6.

Kiskowski, M., & Chowell, G. (2016). Modeling household and community transmission of Ebola virus disease: epidemic growth, spatial dynamics and insights for epidemic control. Virulence, 7(2), 163-173.

Koya, P. R., & Mamo, D. K. (2015). Ebola epidemic disease: modelling, stability analysis, spread control technique, simulation study and data fitting. Journal of Multidisciplinary Engineering Science and Technology, 2(3), 476-484.

Krauer, F., Gsteiger, S., Low, N., Hansen, C. H., & Althaus, C. L. (2016). Heterogeneity in district-level transmission of Ebola virus disease during the 2013-2015 epidemic in West Africa. PLoS neglected tropical diseases, 10(7), e0004867.

Kucharski, A. J., Eggo, R. M., Watson, C. H., Camacho, A., Funk, S., & Edmunds, W. J. (2016). Effectiveness of ring vaccination as control strategy for Ebola virus disease. Emerging infectious diseases, 22(1), 105.

Kullback, S.; Leibler, R. A. (1951). On Information and Sufficiency. Ann. Math. Statist. 22, no. 1, 79--86. doi:10.1214/aoms/1177729694.

Lau, M. S., Dalziel, B. D., Funk, S., McClelland, A., Tiffany, A., Riley, S., ... & Grenfell, B. T. (2017). Spatial and temporal dynamics of superspreading events in the 2014–2015 West Africa Ebola epidemic. Proceedings of the National Academy of Sciences, 201614595.

Lau, M. S., Dalziel, B. D., Funk, S., McClelland, A., Tiffany, A., Riley, S., ... & Grenfell, B. T. (2017). Spatial and temporal dynamics of superspreading events in the 2014â€'2015 West Africa Ebola epidemic. Proceedings of the National Academy of Sciences, 201614595.

Lawrence, P., Danet, N., Reynard, O., Volchkova, V., & Volchkov, V. (2017). Human transmission of Ebola virus. Current opinion in virology, 22, 51-58.

Lewnard, J. A., Mbah, M. L. N., Alfaro-Murillo, J. A., Altice, F. L., Bawo, L., Nyenswah, T. G., & Galvani, A. P. (2014). Dynamics and control of Ebola virus transmission in Montserrado, Liberia: a mathematical modelling analysis. The Lancet Infectious Diseases, 14(12), 1189-1195.

Li, S. L., Bjȩrnstad, O. N., Ferrari, M. J., Mummah, R., Runge, M. C., Fonnesbeck, C. J., ... & Shea, K. (2017). Essential information: Uncertainty and optimal control of Ebola outbreaks. Proceedings of the National Academy of Sciences, 201617482.

Liu, W., Tang, S., & Xiao, Y. (2015). Model selection and evaluation based on emerging infectious disease data sets including A/H1N1 and Ebola. Computational and mathematical methods in medicine, 2015.

Lopes, A. M., Andrade, J. P., & Machado, J. T. (2016). Multidimensional scaling analysis of virus diseases. Computer methods and programs in biomedicine, 131, 97-110.

Lucas, P. J., Van der Gaag, L. C., & Abu-Hanna, A. (2004). Bayesian networks in biomedicine and health-care. Artificial intelligence in medicine, 30(3), 201-214.

MacIntyre, C. R., Chughtai, A. A., Seale, H., Richards, G. A., & Davidson, P. M. (2015). Uncertainty, risk analysis and change for Ebola personal protective equipment guidelines. International journal of nursing studies, 52(5), 899-903.

Majumder, M. S., Kluberg, S., Santillana, M., Mekaru, S., & Brownstein, J. S. (2015). 2014 Ebola outbreak: media events track changes in observed reproductive number. PLoS currents, 7.

Maurer, F. P., Castelberg, C., Von Braun, A., Wolfensberger, A., Bloemberg, G. V., BÃ¶ttger, E. C., & Somoskovi, A. (2014). Postsurgical wound infections due to rapidly growing mycobacteria in Swiss medical tourists following cosmetic surgery in Latin America between 2012 and 2014. Eurosurveillance, 19(37), 20905.

Merler, S., Ajelli, M., Fumanelli, L., Gomes, M. F., y Piontti, A. P., Rossi, L., ... & Vespignani, A. (2015). Spatiotemporal spread of the 2014 outbreak of Ebola virus disease in Liberia and the effectiveness of non-pharmaceutical interventions: a computational modelling analysis. The Lancet Infectious Diseases, 15(2), 204-211.

More, S., Bȩtner, A., Butterworth, A., Calistri, P., Depner, K., Edwards, S., ... & Miranda, M. A. (2017). Assessment of listing and categorisation of animal diseases within the framework of the Animal Health Law (Regulation (EU) No 2016/429): paratuberculosis. Efsa Journal, 15(7).

Morelli, M. J., Thébaud, G., Chadœuf, J., King, D. P., Haydon, D. T., & Soubeyrand, S. (2012). A Bayesian inference framework to reconstruct transmission trees using epidemiological and genetic data. PLoS computational biology, 8(11), e1002768.

Morse, S. S., Mazet, J. A., Woolhouse, M., Parrish, C. R., Carroll, D., Karesh, W. B., ... & Daszak, P. (2012). Prediction and prevention of the next pandemic zoonosis. The Lancet, 380(9857), 1956-1965.

Murphy, K. P. (2002). Dynamic bayesian networks. *Probabilistic Graphical Models, M. Jordan*, 7.

Ngwa, G. A., & Teboh-Ewungkem, M. I. (2016). A mathematical model with quarantine states for the dynamics of ebola virus disease in human populations. Computational and mathematical methods in medicine, 2016.

Nieddu, G. T., Billings, L., Kaufman, J. H., Forgoston, E., & Bianco, S. (2017). Extinction pathways and outbreak vulnerability in a stochastic Ebola model. Journal of The Royal Society Interface, 14(127), 20160847.

Nishiura, H., & Chowell, G. (2015). Theoretical perspectives on the infectiousness of Ebola virus disease. Theoretical Biology and Medical Modelling, 12(1), 1.

Pandey, A., Atkins, K. E., Medlock, J., Wenzel, N., Townsend, J. P., Childs, J. E., ... & Galvani, A. P. (2014). Strategies for containing Ebola in west Africa. Science, 346(6212), 991-995.

Pearl, J. (1988). Probabilistic Reasoning in Intelligent Systems: Networks of Plausible Inference. San Mateo, CA, USA: Morgan Kaufmann Publishers.

Pell, B., Baez, J., Phan, T., Gao, D., Chowell, G., & Kuang, Y. (2016). Patch models of EVD transmission dynamics. In Mathematical and Statistical Modeling for Emerging and Re-emerging Infectious Diseases (pp. 147-167). Springer, Cham.

Pell, B., Kuang, Y., Viboud, C., & Chowell, G. (2016). Using phenomenological models for forecasting the 2015 Ebola challenge. Epidemics.

Perra, N., Gonçalves, B., Pastor-Satorras, R., & Vespignani, A. (2012). Activity driven modeling of time varying networks. Scientific reports, 2, 469.

Sun, W., Park, C. Y., & Carvalho, R. (2011). A New Research Tool for Hybrid Bayesian Networks using Script Language, Proceedings of the SPIE - The International Society for Optical Engineering, Vol. 8050, pp. 80501Q.

Poletto, C., Gomes, M. F., y Piontti, A. P., Rossi, L., Bioglio, L., Chao, D. L., ... & Vespignani, A. (2014). Assessing the impact of travel restrictions on international spread of the 2014 West African Ebola epidemic. Euro surveillance: bulletin Europeen sur les maladies transmissibles= European communicable disease bulletin, 19(42).

Pruyt, E., Auping, W. L., & Kwakkel, J. H. (2015). Ebola in West Africa: Model‐Based Exploration of Social Psychological Effects and Interventions. Systems Research and Behavioral Science, 32(1), 2-14.

Rachah, A., & Torres, D. F. (2015). Mathematical modelling, simulation, and optimal control of the 2014 Ebola outbreak in West Africa. Discrete Dynamics in Nature and Society, 2015.

Rachah, A., & Torres, D. F. (2016). Dynamics and optimal control of Ebola transmission. Mathematics in Computer Science, 10(3), 331-342.

Rivers, C. M., Lofgren, E. T., Marathe, M., Eubank, S., & Lewis, B. L. (2014). Modeling the impact of interventions on an epidemic of Ebola in Sierra Leone and Liberia. PLoS currents, 6.

Rizzo, A., Pedalino, B., & Porfiri, M. (2016). A network model for Ebola spreading. Journal of theoretical biology, 394, 212-222.

Rosello, A., Mossoko, M., Flasche, S., Van Hoek, A. J., Mbala, P., Camacho, A., ... & Piot, P. (2015). Ebola virus disease in the Democratic Republic of the Congo, 1976-2014. Elife, 4, e09015.

Scarpino, S. V., Iamarino, A., Wells, C., Yamin, D., Ndeffo-Mbah, M., Wenzel, N. S., ... & Meyers, L. A. (2014). Epidemiological and viral genomic sequence analysis of the 2014 Ebola outbreak reveals clustered transmission. Clinical Infectious Diseases, 60(7), 1079-1082.

Sharma, N., & Cappell, M. S. (2015). Gastrointestinal and hepatic manifestations of Ebola virus infection. Digestive diseases and sciences, 60(9), 2590-2603.

Shen, M., Xiao, Y., & Rong, L. (2015). Modeling the effect of comprehensive interventions on Ebola virus transmission. Scientific reports, 5, 15818.

Shwe, M. A., Middleton, B., Heckerman, D. E., Henrion, M., Horvitz, E. J., Lehmann, H. P., & Cooper, G. F. (1991). Probabilistic diagnosis using a reformulation of the INTERNIST-1/QMR knowledge base. Methods of information in Medicine, 30(4), 241-255.

Siettos, C. I., & Russo, L. (2013). Mathematical modeling of infectious disease dynamics. Virulence, 4(4), 295-306.

Siettos, C., Anastassopoulou, C., Russo, L., Grigoras, C., & Mylonakis, E. (2015). Modeling the 2014 ebola virus epidemic–agent-based simulations, temporal analysis and future predictions for liberia and sierra leone. PLoS currents, 7.

Singleton, K. W. (2016). Investigating Predictive Disease Model Transportability through Cohort Simulation and Causal Analysis (Doctoral dissertation, UCLA).

Stadler, T., Kühnert, D., Rasmussen, D. A., & du Plessis, L. (2014). Insights into the early epidemic spread of Ebola in Sierra Leone provided by viral sequence data. PLoS currents, 6.

Taylor, B. P., Dushoff, J., & Weitz, J. S. (2016). Stochasticity and the limits to confidence when estimating R0 of Ebola and other emerging infectious diseases. Journal of theoretical biology, 408, 145-154.

Thompson, R. N., Gilligan, C. A., & Cunniffe, N. J. (2016). Detecting presymptomatic infection is necessary to forecast major epidemics in the earliest stages of infectious disease outbreaks. PLoS computational biology, 12(4), e1004836.

Towers, S., Patterson-Lomba, O., & Castillo-Chavez, C. (2014). Temporal variations in the effective reproduction number of the 2014 West Africa Ebola outbreak. PLoS currents, 6.

Troncoso, A. (2015). Ebola outbreak in West Africa: a neglected tropical disease. Asian Pacific Journal of Tropical Biomedicine, 5(4), 255-259.

van den Driessche, P. (2017). Reproduction numbers of infectious disease models. Infectious Disease Modelling, 2(3), 288-303.

Van Kerkhove, M. D., Bento, A. I., Mills, H. L., Ferguson, N. M., & Donnelly, C. A. (2015). A review of epidemiological parameters from Ebola outbreaks to inform early public health decision-making. Scientific data, 2, 150019.

Velikova, M. V., Terwisscha van Scheltinga, J. A., Lucas, P. J., & Spaanderman, M. (2014). Exploiting causal functional relationships in Bayesian network modelling for personalised healthcare.

Volz, E., & Pond, S. (2014). Phylodynamic analysis of Ebola virus in the 2014 Sierra Leone epidemic. PLoS currents, 6.

Wallinga, J., & Lipsitch, M. (2007). How generation intervals shape the relationship between growth rates and reproductive numbers. Proceedings of the Royal Society of London B: Biological Sciences, 274(1609), 599-604.

Webb, G. F., & Browne, C. J. (2016). A model of the Ebola epidemics in West Africa incorporating age of infection. Journal of biological dynamics, 10(1), 18-30.

Webb, G., Browne, C., Huo, X., Seydi, O., Seydi, M., & Magal, P. (2015). A model of the 2014 Ebola epidemic in West Africa with contact tracing. PLoS currents, 7.

Weitz, J. S., & Dushoff, J. (2015). Modeling post-death transmission of Ebola: Challenges for inference and opportunities for control. Scientific reports, 5, 8751.

WHO Ebola Response Team. (2014). Ebola virus disease in West Africa—the first 9 months of the epidemic and forward projections. New England Journal of Medicine, 371(16), 1481-1495.

Wong, J. Y., Zhang, W., Kargbo, D., Haque, U., Hu, W., Wu, P., ... & Yang, R. (2016). Assessment of the severity of Ebola virus disease in Sierra Leone in 2014–2015. Epidemiology & Infection, 144(7), 1473-1481.

Wong, Z. S. Y., Bui, C. M., Chughtai, A. A., & Macintyre, C. R. (2017). A systematic review of early modelling studies of Ebola virus disease in West Africa. Epidemiology & Infection, 145(6), 1069-1094.

Xia, Z. Q., Wang, S. F., Li, S. L., Huang, L. Y., Zhang, W. Y., Sun, G. Q., ... & Jin, Z. (2015). Modeling the transmission dynamics of Ebola virus disease in Liberia. Scientific reports, 5, 13857.

Yamin, D., Gertler, S., Ndeffo-Mbah, M. L., Skrip, L. A., Fallah, M., Nyenswah, T. G., ... & Galvani, A. P. (2015). Effect of Ebola progression on transmission and control in Liberia. Annals of internal medicine, 162(1), 11-17.

Zakary, O., Rachik, M., & Elmouki, I. (2017). A multi-regional epidemic model for controlling the spread of Ebola: awareness, treatment, and travel-blocking optimal control approaches. Mathematical Methods in the Applied Sciences, 40(4), 1265-1279.

Zarandi, M. F., Seifi, A., Ershadi, M. M., & Esmaeeli, H. (2017, October). An Expert System Based on Fuzzy Bayesian Network for Heart Disease Diagnosis. In North American Fuzzy Information Processing Society Annual Conference (pp. 191-201). Springer, Cham.

Ebloa Map Data. (2018). locations.geojson https://github.com/cmrivers/ebola.